\pdfoutput=1

\documentclass[11pt]{article}

\usepackage[]{ACL2023}

\usepackage{times}
\usepackage{latexsym}
\usepackage[caption=false]{subfig}

\usepackage[T1]{fontenc}

\usepackage[utf8]{inputenc}

\usepackage{microtype}
\usepackage{graphicx}
\usepackage{amsmath}
\usepackage{tikz}
\usepackage{multirow}
\usepackage{overpic}

\usepackage{amssymb}
\usepackage{pifont}

\usepackage{inconsolata}

\title{VLN-Trans: Translator for the Vision and Language Navigation Agent}

\author{Yue Zhang \\
  Michigan State University \\
  \texttt{zhan1624@msu.edu} \\\And
  Parisa Kordjamshidi \\
  Michigan State University \\
  \texttt{kordjams@msu.edu} \\}

\begin{document}
\maketitle
\begin{abstract}
Language understanding is essential for the navigation agent to follow instructions. We observe two kinds of issues in the instructions that can make the navigation task challenging: 
1. The mentioned landmarks are not recognizable by the navigation agent due to the different vision abilities of the instructor and the modeled agent. 2. The mentioned landmarks are applicable to multiple targets, thus not distinctive for selecting the target among the candidate viewpoints.
To deal with these issues, we design a translator module for the navigation agent to convert the original instructions into easy-to-follow sub-instruction representations at each step. The translator needs to focus on the recognizable and distinctive landmarks based on the agent's visual abilities and the observed visual environment.
To achieve this goal, we create a new synthetic sub-instruction dataset and design specific tasks to train the translator and the navigation agent.
We evaluate our approach on Room2Room~(R2R), Room4room~(R4R), and Room2Room Last (R2R-Last) datasets and achieve state-of-the-art results on multiple benchmarks.
\end{abstract}
\section{Introduction}
Vision-and-Language Navigation~(VLN)~\cite{anderson2018vision} task requires an agent to understand and follow complex instructions to arrive at a destination in a photo-realistic simulated environment. This cross-domain task attracts researchers from the communities of computer vision, natural language processing, and robotics~\cite{gu2022vision, wu2021vision,francis2022core}.
\begin{figure}
    \centering
    \includegraphics[width=1.0\linewidth]{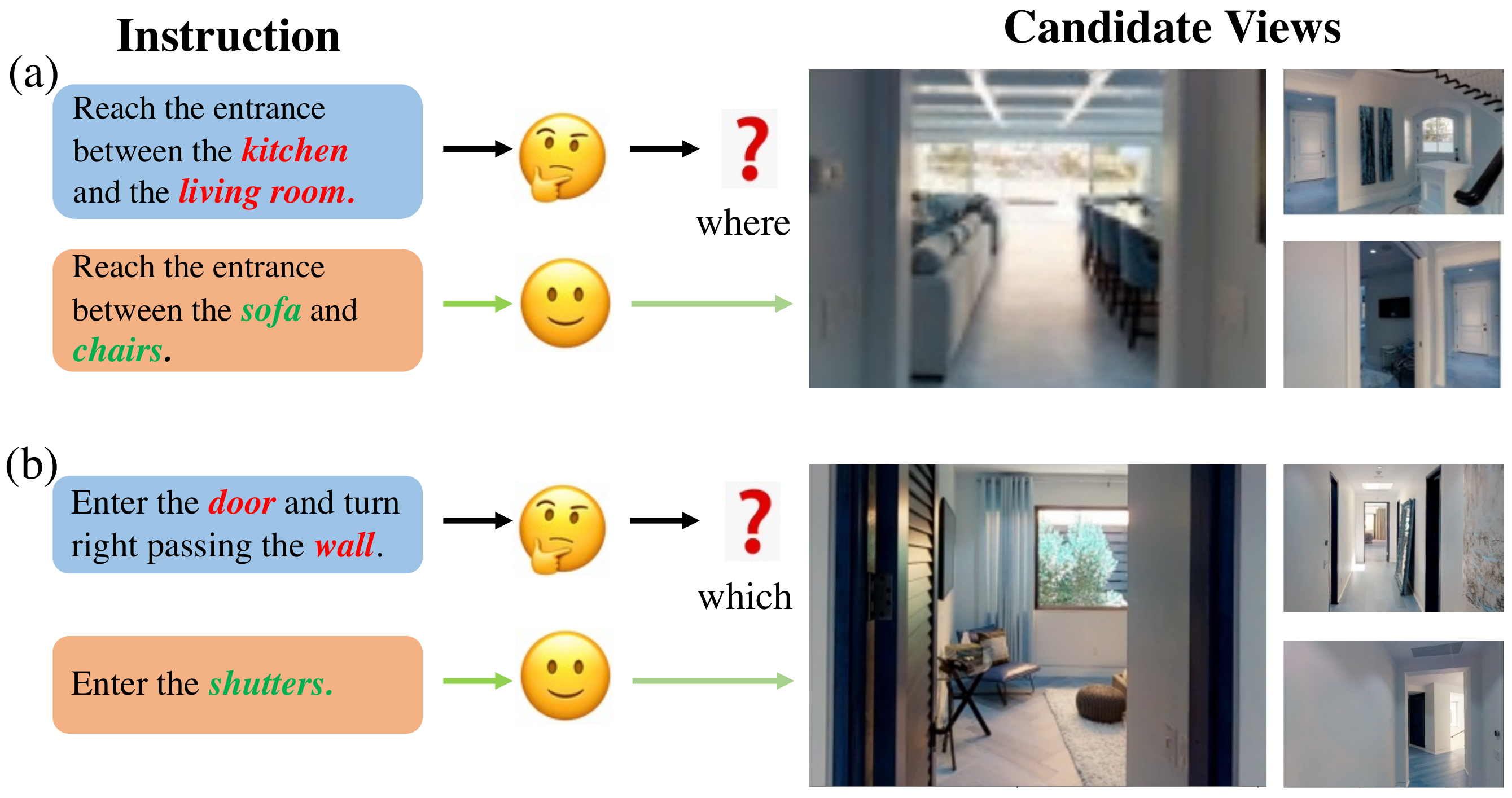}
    \caption{Two types of instructions that make the grounding in the VLN task challenging: (a) unrecognizable landmarks, (b) nondistinctive landmarks. In candidate views, the largest image shows the target view.}
    \vspace{-5mm}
    \label{fig:case example1}
\end{figure}

To solve the VLN task, one streamline of methods is to build the connections between text and vision modalities by grounding the semantic information dynamically~\cite{hong2020language,qi2020object,an2021neighbor, zhang2022explicit}.
However, we observe two types of instructions that make the grounding in the VLN task quite challenging. \textbf{First}, the instruction contains landmarks that are not recognizable by the navigation agent.
For example, Figure~\ref{fig:case example1}(a), the agent can only see the ``sofa'', ``table'' and ``chair'' in the target viewpoint, based on the learned vision representations~\cite{he2016deep,ren2015faster,dosovitskiy2020image}.
However, the instructor mentions landmarks of the ``living room'' and ``kitchen'' in the instruction, based on their prior knowledge about the environment, such as relating ``sofa'' to ``living room''.
Given the small size of the dataset designed for learning navigation, it is hard to expect the agent to gain the same prior knowledge as the instructor.
\textbf{Second}, the instructions contain the landmarks that can be applied to multiple targets, which causes ambiguity for the navigating agent. In Figure~\ref{fig:case example1}(b), the instruction ``enter the door'' does not help distinguish the target viewpoint from other candidate viewpoints since there are multiple doors and walls in the visual environment. 
As a result, we hypothesize those types of instructions cause the explicit and fine-grained grounding to be less effective for the VLN task, as appears in~\cite{hong2020sub, zhang2021towards} that use sub-instructions and in~\cite{hong2020language, hu2019you, qi2020object, zhang2022explicit} that use object-level representations. 

To address the aforementioned issues, the main idea in our work is to introduce a translator module in the VLN agent, named VLN-trans, which takes the given instruction and visual environment as inputs and then converts them to easy-to-follow sub-instructions focusing on two aspects: 1) \textit{recognizable} landmarks based on the navigation agent's visualization ability. 2) \textit{distinctive} landmarks that help the navigation agent distinguish the targeted viewpoint from the candidate viewpoints. 
Consequently, by focusing on those two aspects, the translator can enhance the connections between the given instructions and the agent's observed visual environment and improve the agent's navigation performance.

To train the translator module, we propose a Synthetic Fine-grained Sub-instruction dataset called SyFiS. The SyFiS dataset consists of pairs of the sub-instructions and their corresponding viewpoints, and each sub-instruction contains a motion indicator and a landmark. 
We select a motion verb for an action based on our action definitions according to the relative directions between source and target viewpoints; To obtain the landmarks, we first use Contrastive Language-Image Pretraining~(CLIP)~\cite{radford2021learning},
a vision \& language pre-trained model with powerful cross-modal alignment ability, to detect the objects in each candidate viewpoint as the recognizable landmarks. Then we select the distinctive one among recognizable landmarks that only appears in the target viewpoint. We train the translator in a contrastive manner by designing positive and negative sub-instructions based on whether a sub-instruction contains distinctive landmarks.

We design two tasks to pre-train the translator: \textit{Sub-instruction Generation} (SG) and \textit{Distinctive Sub-instruction Learning} (DSL). The SG task enables the translator to generate the correct sub-instruction. The DSL task encourages the translator to learn effective sub-instruction representations that are close to positive sub-instructions with distinctive landmarks and are far from the negative sub-instructions with irrelevant and nondistinctive landmarks. Then we equip the navigation agent with the pre-trained translator. At each navigation step, the translator adaptively generates easy-to-follow sub-instruction representations for the navigation agent based on given instructions and the agent's current visual observations. During the navigation process, we further design an auxiliary task, \textit{Sub-instruction Split} (SS), to optimize the translator module to focus on the important portion of the given instruction and generate more effective sub-instruction representations. 

\noindent In summary, our contributions are as follows:

1. We propose a translator module that helps the navigation agent generate easy-to-follow sub-instructions considering recognizable and distinctive landmarks based on the agent's visual ability. 

2. We construct a high-quality synthetic sub-instruction dataset and design specific tasks for training the translator and the navigation agent.

3. We evaluate our method on R2R, R4R, and R2R-Last, and our method achieves the SOTA results on all benchmarks.

\section{Related Work}
\textbf{Vision-and-Language Navigation} 
\newcite{anderson2018vision} first propose the VLN task with R2R dataset, and many LSTM-based models~\cite{tan2019learning, ma2019self, wang2019reinforced, ma2019regretful} show progressing performance. One line of research on this task is to improve the grounding ability by modeling the semantic structure of both the text and vision modalities~\cite{hong2020language,li2021improving,zhang2022explicit}. 
Recently, Transformers~\cite{vaswani2017attention, tan2019lxmert,hong2021vln} have been broadly used in the VLN task. 
VLN$\circlearrowright$BERT~\cite{hong2021vln} equips a Vision and Language Transformer with a recurrent unit that uses the history information, and HAMT~\cite{chen2021history} has an explicit history learning module and uses Vision Transformer~\cite{dosovitskiy2020image} to learn vision representations.  
To improve learning representation for the agent, ADAPT~\cite{lin2022adapt} learns extra prompt features, and CITL~\cite{liang2022contrastive} proposes a contrastive instruction-trajectory learning framework.
However, previous works ignore the issue of unrecognizable and nondistinctive landmarks in the instruction, which is detrimental to improving the navigation agent's grounding ability. We propose a translator module that generates easy-to-follow sub-instructions, which helps the agent overcome the abovementioned issues and improves the agent's navigation performance.\\
\textbf{Instruction Generation} \newcite{fried2018speaker} propose an instruction generator (\textit{e.g.}, Speaker) to generate instructions as the offline augmented data for the navigation agent. \newcite{kurita2020generative} design a generative language-grounded policy for the VLN agent to compute the distribution over all possible instructions given action and transition history. Recently, FOAM~\cite{dou2022foam} uses a bi-level learning framework to model interactions between the navigation agent and the instruction generator. \newcite{wang2022counterfactual} propose a cycle-consistent learning scheme that learns both instruction following and generation tasks. 
{In contrast to our work, most prior works rely on the entire trajectory to generate instructions that provide a rather weak supervision signal for each navigation action. Moreover, the previously designed speakers generate textual tokens based on a set of images without considering what instructions are easier for the agent to follow.
We address those issues with our designed translator by generating easy-to-follow sub-instruction representations for the navigation agent at each navigation step based on recognizable and distinctive landmarks.}
\vspace{-3mm}
\section{Method}
\begin{figure}[t]
    \centering
    \begin{overpic}[width=1.0\linewidth]{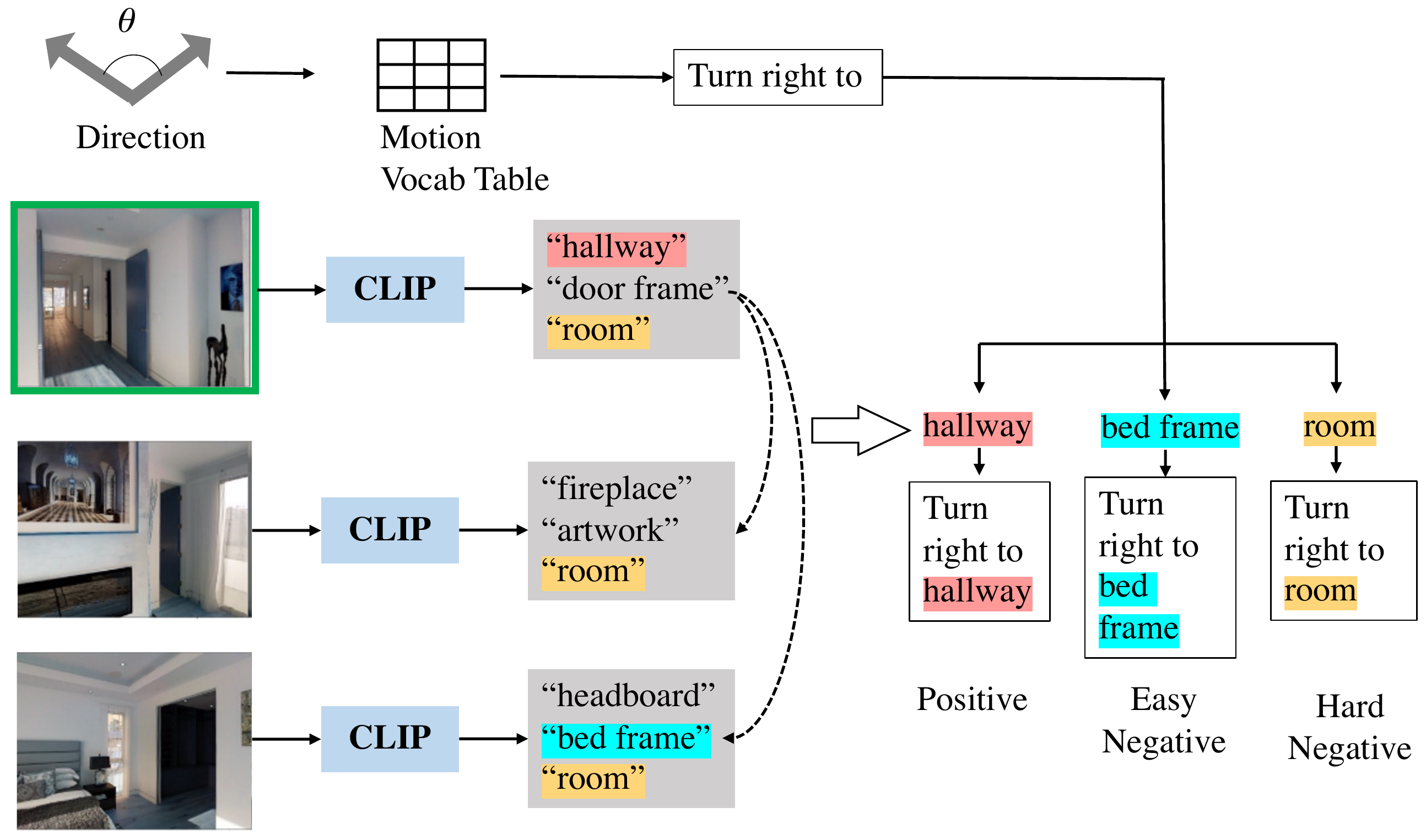}
    \put(5,29){\scriptsize{v1 (target)}}
    \put(5, 13){\scriptsize{v2}}
    \put(5,-2){\scriptsize{v3}}
    \end{overpic}
    \caption{Illustration of constructing the SyFiS dataset.
    }
    \vspace{-5mm}
    \label{fig:synthetic data}
\end{figure}

In our navigation problem setting, the agent is given an instruction, denoted as $W=\{w_1, w_2, \cdots, w_L\}$, where $L$ is the number of tokens. Also, the agent observes a panoramic view including $36$ viewpoints\footnote{$12$ headings and $3$ elevations with $30$ degree interval.} at each navigation step. There are $n$ candidate viewpoints that the agent can navigate to in a panoramic view, denoted as $I=\{I_{1}, I_{2}, \cdots, I_{n}\}$. The task is to generate a trajectory that takes the agent close to a goal destination. 
The navigation terminates when the navigation agent selects the current viewpoint, or a pre-defined maximum navigation step is reached.

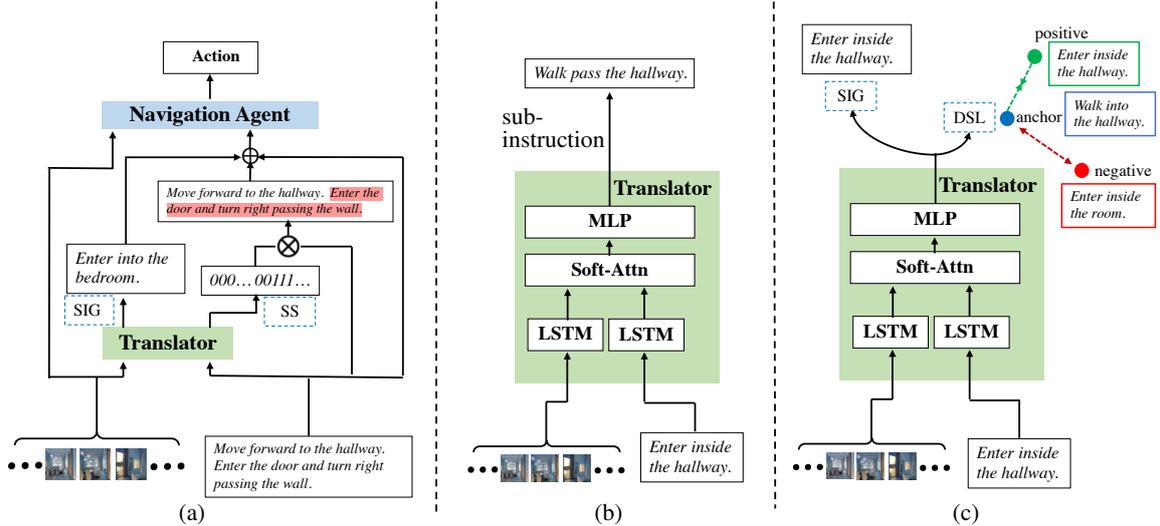
\begin{figure*}
    \centering
    \begin{overpic}[width=0.95\linewidth]{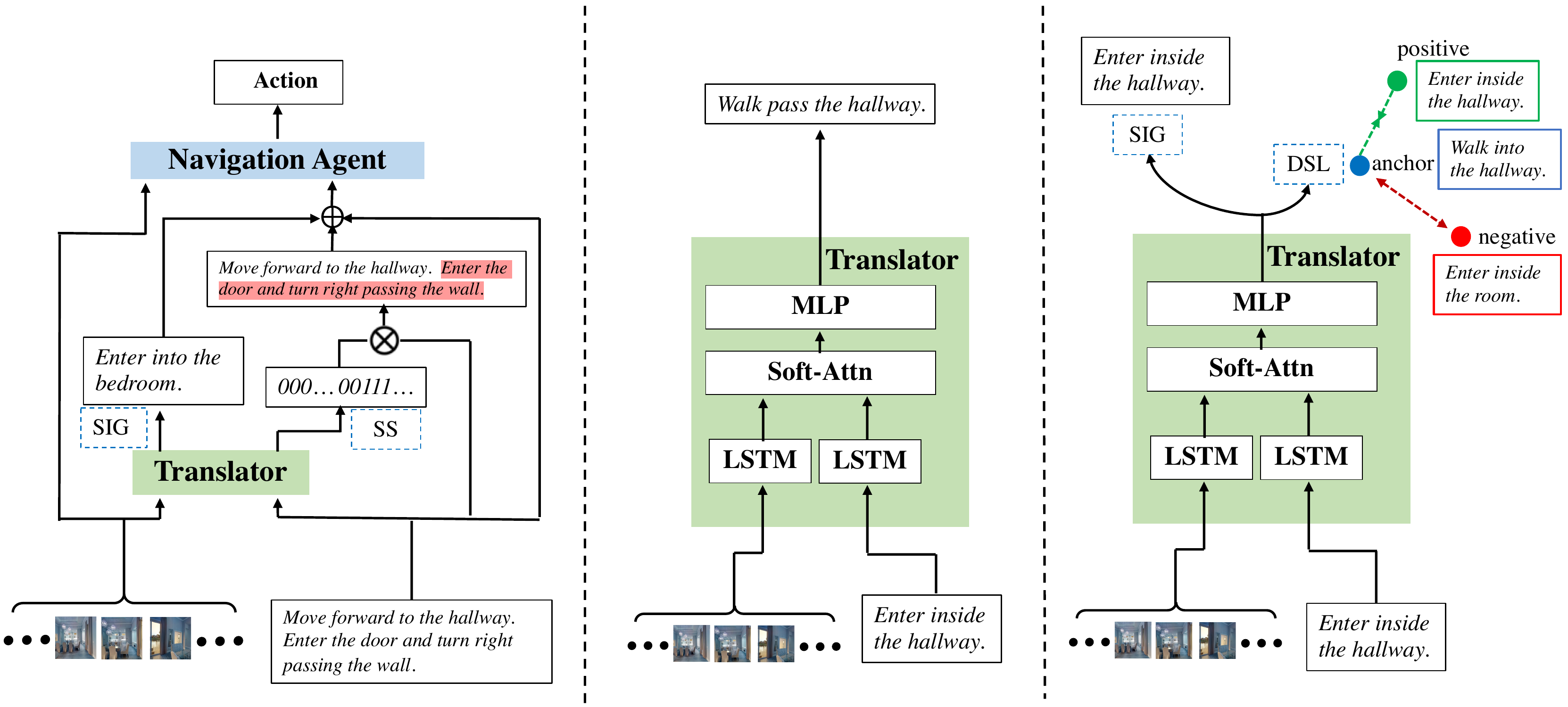}
    \put(43,34){\small{sub-}}
    \put(43,32){\small{instruction}}
    
    \put(15,-0.5){\small{(a)}}
    \put(51,-0.5){\small{(b)}}
    \put(82,-0.5){\small{(c)}}
    \end{overpic}
    \caption{The overview of the proposed method. (a) Navigation agent with VLN-Trans. (b) The translator architecture (c) Pre-training the translator. SIG:Sub-instruction Generation; DSL: Distinctive Sub-instruction Learning; SS: Sub-instruction Split.
    }
    \vspace{-3mm}
    \label{fig:Navigation Agent}
\end{figure*}

Fig.~\ref{fig:Navigation Agent}~(a) provides an overall picture of our proposed architecture for the navigation agent. We use VLN$\circlearrowright$BERT~\cite{hong2021vln} (in Sec.~\ref{Navigation Agent}) as the backbone of our navigation agent and equip it with a novel \textit{translator} module that is trained to convert the full instruction representation into the most relevant sub-instruction representation based on the current visual environment.
Another key point of our method is to create a synthetic sub-instruction dataset and design the pre-training tasks to encourage the translator to generate effective sub-instruction representations. We describe the details of our method in the following sections.

\subsection{Backbone: VLN$\circlearrowright$BERT}
\label{Navigation Agent}
We use VLN$\circlearrowright$BERT as the backbone of our navigation agent. It is a cross-modal Transformer-based navigation agent with a specially designed recurrent state unit. 
At each navigation step, the agent takes three inputs: text representation,
vision representation, and state representation.
The text representation $X$ for instruction $W$ is denoted as $X=[x_1, x_2, \cdots, x_L]$.
The vision representation $V$ for candidate viewpoints $I$ is denoted as $V=[v_1, v_2, \cdots, v_n]$. The recurrent state representation $S_t$ stores the history information of previous steps and is updated based on $X$ and $V_t$ at the current step.
The state representation $S_t$ along with $X$ and $V_t$ are passed to cross-modal transformer layers and self-attention layers to learn the cross-modal representations and select an action, as follows:
\begin{equation}
\label{main equation}
    \hat{X}, \hat{S_t}, \hat{V_t} = Cross\_Attn(X, [S_t;V_t]),
\end{equation}
\vspace{-6mm}
\begin{equation}
    S_{t+1}, a_t = Self\_Attn(\hat{S_t}, \hat{V_t}),
\end{equation}
we use $\hat{X}$, $\hat{S_t}$, $\hat{V_t}$ to represent text, recurrent state, and visual representations after cross-modal transformer layers, respectively. The action is selected based on the self-attention scores between $\hat{S_t}$ and $\hat{V_t}$. $S_{t+1}$ is the updated state representations and  $a_t$ contains the probability of the actions. 

\subsection{Synthetic Sub-instruction Dataset (SyFiS)}
\label{Synthetic Data}
This section introduces our novel approach to automatically generate a synthetic fine-grained sub-instruction dataset, SyFiS, which is used to pre-train the \textit{translator} (described in Sec.~\ref{VLN-Trans}) in a contrastive manner.
To this aim, for each viewpoint, we generate one positive sub-instruction and three negative sub-instructions. The viewpoints are taken from the R2R dataset~\cite{anderson2018vision}, and the sub-instructions are generated based on our designed template. Fig.~\ref{fig:synthetic data} shows an example describing our methodology for constructing the dataset.  The detailed statistics of our dataset are included in Sec.\ref{qualitative study}.

The sub-instruction template includes two components: a motion indicator and a landmark. For example, in the sub-instruction ``turn left to the kitchen'', the motion indicator is ``turn left'', and the landmark is ``kitchen''. The sub-instruction template is designed based on the semantics of \textit{Spatial Configurations} explained in~\cite{dan2020spatial}. 

\noindent\textbf{Motion Indicator Selection}
First, we generate the motion indicator for the synthesized sub-instructions.
Following \newcite{zhang2021towards}, we use pos-tagging information to extract the verbs from instructions in the R2R training dataset and form our motion-indicators dictionary. 
We divide the motion indicators to $6$ categories of:  ``FORWARD'', ``LEFT'', ``RIGHT'', ``UP'', ``DOWN'', and ``STOP''. Each category has a set of corresponding verb phrases.
We refer the Appendix~\ref{motion indicator dictionary} for more details about motion indicator dictionary.

Given a viewpoint, to select a motion indicator for each sub-instruction, we calculate the differences between the elevation and headings of the current and the target viewpoints. Based on the orientation difference and a threshold, \textit{e.g.} $30$ degrees, we decide the motion-indicator category. 
Then we randomly pick a motion verb from the corresponding category to be used in both generated positive and negative sub-instructions. 
\\
\textbf{Landmark Selection} 
For generating the landmarks for the sub-instructions, we use the candidate viewpoints at each navigation step and select the most \textit{recognizable} and \textit{distinctive} landmarks that are easy for the navigation agent to follow. 

In our approach, the most recognizable landmarks are the objects that can be detected by CLIP. Using CLIP~\cite{radford2021learning}, given a viewpoint image,  we predict a {label} token with the prompt ``a photo of {label}'' from an object label vocabulary. The probability that the image with representation $b$ contains a label $c$ is calculated as follows,
\begin{equation}
    p(c) = \frac{exp(sim(b, w_c)/\tau_1)}{\sum^M_{i=1}(exp(sim(b, w_i))/\tau_1)},
    \label{eq_pro_cal}
\end{equation}
\noindent where $\tau_1$ is the temperature parameter, $sim$ is the cosine similarity between image representation and phrase representation $w_c$ which are generated by CLIP~\cite{radford2021learning}, $M$ is the vocabulary size. 
The top-$k$ objects that have the maximum similarity with the image are selected to form the set of recognizable landmarks for each viewpoint.

We filter out the distinctive landmarks from the recognizable landmarks. 
The distinctive landmarks are the ones that appear in the target viewpoint and not in any other candidate viewpoints.
For instance, in the example of Fig.~\ref{fig:synthetic data}, ``hallway'' is a distinctive landmark because it only appears in the v1 (target viewpoint).

\noindent\textbf{Forming Sub-instructions} We use the motion verbs and landmarks to construct sub-instructions based on our template. 
To form contrastive learning examples, we create positive and negative sub-instructions for each viewpoint. A positive sub-instruction is a sub-instruction that includes a distinctive landmark. 
The negative sub-instructions include easy negatives and hard negatives. 
An easy negative sub-instruction contains irrelevant landmarks that appear in any candidate viewpoint except the target viewpoint, \textit{e.g.,} in Fig.~\ref{fig:synthetic data}, ``bed frame'' appears in v3 and is not observed in the target viewpoint.
A hard negative sub-instruction includes the nondistinctive landmarks that appear in both the target viewpoint and other candidate viewpoints. For example,  in Fig.~\ref{fig:synthetic data}, ``room'' can be observed in all candidate viewpoints; therefore, it is difficult to distinguish the target from other candidate viewpoints based on this landmark. 

\subsection{Translator Module}
\label{VLN-Trans}
The translator takes a set of candidate viewpoints and the corresponding sub-instruction as the inputs and generates new sub-instructions. 
The architecture of our translator is shown in Fig.~\ref{fig:Navigation Agent}(b). This architecture is similar to the LSTM-based Speaker in the previous works~\cite{tan2019learning, fried2018speaker}. However,
they generate full instructions from the whole trajectories and use them as offline augmented data for training the navigation agent, while our translator adaptively generates sub-instruction during the agent's navigation process based on its observations at each step.

Formally, we feed text representations of sub-instruction $X$ and the visual representations of candidate viewpoints $V$ into the corresponding LSTM to obtain deeper representation $\tilde{X}$ and $\tilde{V}$. Then, we apply the soft attention between them to obtain the visually attended text representation $\tilde{X}'$, as:
\begin{equation}
\small
\label{weight equation}
    \tilde{X}'= SoftAttn(\tilde{X}; \tilde{V}; \tilde{V})=
    softmax(\tilde{X}^TW\tilde{V})\tilde{V},
\end{equation}
where $W$ is the learned weights. Lastly, we use an MLP layer to generate sub-instruction $X'$ from the hidden representation $\tilde{X}'$, as follows,
\begin{equation}
    X' = softmax(MLP(\tilde{X}'))
\end{equation}

We use the SyFiS dataset to pre-train this translator. We also design two pre-training tasks: Sub-instruction Generation and Distinctive sub-instruction Learning. \\
\textbf{Sub-instruction Generation (SIG)}  We first train the translator to generate a sub-instruction, given the positive instructions paired with the viewpoints in the SyfiS dataset as the ground-truth. We apply a cross-entropy loss between the generated sub-instruction $X'$ and the positive sub-instruction $X_p$. The loss function for the SIG task is as follows,
\begin{equation}
    L_{SIG} = -\frac{1}{L} \sum_L X_p log P(X')
\end{equation}
\noindent\textbf{Distinctive Sub-instruction Learning (DSL)}
To encourage the translator to learn sub-instruction representations that are close to the positive sub-instructions with recognizable and distinctive landmarks, and are far from the negative sub-instructions with irrelevant and nondistinctive landmarks, we use triplet loss to train the translator in a contrastive way. To this aim, we first design triplets of sub-instructions in the form of <anchor, positive, negative>. For each viewpoint, we select one positive and three negative sub-instructions forming three triplets per viewpoint.
We obtain the anchor sub-instruction by replacing the motion indicator in the positive sub-instruction with a different motion verb in the same motion indicator category. 
We denote the text representation of anchor sub-instruction as $X_a$, positive sub-instruction as $X_p$, and negative sub-instruction as $X_n$. Then we feed them to the translator to obtain the corresponding hidden representations $\tilde{X}_a'$, $\tilde{X}_p'$, and $\tilde{X}_n'$ using Eq.~\ref{weight equation}.
The triplet loss function for the DSL task is computed as follows,
\begin{equation}
    L_{DSL} = max(D(\tilde{X}_a', \tilde{X}_p')-D(X',\tilde{X}_n')+m, 0),
\end{equation}
where $m$ is a margin value to keep negative samples far apart, $D$ is the pair-wise distance between representations. In summary, the total objective to pre-train the translator is:
\begin{equation}
    L_{pre-train} = \alpha_1L_{SIG} + \alpha_2L_{DSL}
    \label{Eq_pretrain}
\end{equation}
where $\alpha_1$ and $\alpha_2$ are hyper-parameters for balancing the importance of the two losses.



\subsection{Navigation Agent}
We place the pre-trained translator module on top of the backbone navigation agent to perform the navigation task. Fig.\ref{fig:Navigation Agent}(a) shows the architecture of our navigation agent.

\subsubsection{VLN-Trans: VLN with Translator}

At each navigation step, the translator takes the given instruction and the current candidate viewpoints as input and generates new sub-instruction representations, which are then used as an additional input to the navigation agent.

Since the given instructions describe the full trajectory, we enable the translator module to focus on the part of the instruction that is in effect at each step.
To this aim, we design another MLP layer in the translator to map the hidden states to a scalar attention representation. Then we do the element-wise multiplication between the attention representation and the instruction representation to obtain the attended instruction representation.

In summary, we first input the text representation of given instruction $X$ and visual representation of candidate viewpoints $V$ to the translator to obtain the translated sub-instruction representation $\tilde{X}'$ using Eq.~\ref{weight equation}.
Then we input  $\tilde{X}'$ to another MLP layer to obtain the attention representation $X_m'$, $ X_m' = MLP(\tilde{X}')$. Then we obtain the attended sub-instruction representation as $X'' = X_m'\odot{X}$, where $\odot$ is the element-wise multiplication. 

Lastly, we input text representation $X$ along with translated sub-instruction representation $\tilde{X}'$ and the attended instruction representation  $X''$ into the navigation agent. In such a case, we update the text representation $X$ of VLN$\circlearrowright$BERT as $[X;\tilde{X}';X'']$, where $;$ is the concatenation operation.


\subsubsection{Training and Inference}
\label{train_and_infer}

We follow~\cite{tan2019learning} to train our navigation agent with a mixture of Imitation Learning (IL) and Reinforcement Learning (RL). The IL is to minimize the cross-entropy loss of the predicted and the ground-truth actions.
RL is to sample an action from the action probability to learn the rewards. The navigation objective is denoted as:
\begin{equation}
\label{equation:nav}
    L_{nav} = -\sum_t{-a^{s}_{t}log(p^a_t)} - \lambda \sum_ta^*_t log(p^a_t)
    \vspace{-3mm}
\end{equation}
where $a^s_t$ is the sampled action for RL, $a^*_t$ is the teacher action, and $\lambda$ is the coefficient.

During the navigation process, we design two auxiliary tasks specific to the translator. The first task is still the SIG task in pre-training to generate the correct sub-instructions; the second task is Sub-instruction Split (SS), which generates the correct attended sub-instruction. 
Specifically, for the SS task, at each step, we obtain the ground-truth attention representation by labeling the tokens of the sub-instruction in the full instruction as $1$ and other tokens as $0$. We denote ground-truth attended sub-instruction representation as $X_m$. Then, we apply Binary Cross Entropy loss between $X_m$ and the generated attention representation $X_m'$ as follows,
\begin{equation}
    L_{SS} =  -\frac{1}{L}\sum_LX_m log(X_m')
\end{equation}
The overall training objective of the navigation agent including the translator's auxiliary tasks is:
\begin{equation}
    L_{obj} = \beta_1 L_{nav} + \beta_2 L_{SIG} + \beta_3L_{SS},
    \label{Eq_finetune}
\end{equation}
where $ \beta_1$,  $\beta_1$, and  $\beta_3$ are the coefficients. 
During inference, we use the greedy search to select an action with the highest probability at each step to finally generate a trajectory.

\section{Experiments}
\begin{table*}[t]
\centering
\small
    \begin{center}
    \resizebox{1.0\textwidth}{!}{
    \begin{tabular}{c c c c c c c c c c c}
    \hline
      &  & \multicolumn{3}{c}{\textbf{Val seen}} & \multicolumn{3}{c}{\textbf{Val Unseen}} & \multicolumn{3}{c}{\textbf{Test Unseen}}\\
    \hline
        & Method & \textbf{NE} $\downarrow$ & \textbf{SR} $\uparrow$ & \textbf{SPL}$\uparrow$ & \textbf{NE} $\downarrow$ &  \textbf{SR} $\uparrow$ & \textbf{SPL}$\uparrow$ & \textbf{NE} $\downarrow$&
         \textbf{SR} $\uparrow$& \textbf{SPL} $\uparrow$\\
    \hline
    $1$ & Env-Drop~\cite{tan2019learning}  & $3.99$ & $0.62$ & $0.59$ & $5.22$  & $0.47$ & $0.43$ & $5.23$ & $0.51$ & $0.47$ \\
    $2$ & RelGraph~\cite{hong2020language}~& $3.47$ &  $0.67$ & $0.65$ & $4.73$ & $0.57$ & $0.53$ & $4.75$ & $0.55$ & $0.52$ \\
    $3$ & NvEM~\cite{an2021neighbor} & $3.44$ & $0.69$ & $0.65$ & $4.27$ & $0.60$ & $0.55$ & $4.37$ & $0.58$ & $0.54$ \\
    \hline
    $4$ & PREVALENT~\cite{hao2020towards} & $3.67$ & $0.69$ & $0.65$ &  $4.71$ &  $0.58$ & $0.53$ & $5.30$ & $0.54$ & $0.51$\\
    $5$ & HAMT (ResNet) ~\cite{chen2021history} & $-$ & $0.69$ & $0.65$ & $-$ & $0.64$ & $0.58$ & $-$ & $-$ & $-$ \\
    $6$ & HAMT (ViT) ~\cite{chen2021history} & $2.51$ & $0.76$ & $0.72$ & $-$ & $0.66$ & $0.61$ & $\mathbf{3.93}$ & $0.65$ & $\mathbf{0.60}$ \\
    $7$ & CITL~\cite{liang2022contrastive} & $2.65$ & $0.75$ & $0.70$ & $3.87$ & $0.63$ & $0.58$ & $3.94$ & $0.64$ & $0.59$\\
    $8$ & ADAPT~\cite{lin2022adapt} & $2.70$ & $0.74$ & $0.69$ & $3.66$ & $0.66$ & $0.59$ & $4.11$ & $0.63$ & $0.57$\\
    $9$ & LOViS~\cite{zhang2022lovis} & $\mathbf{2.40}$ & $0.77$ & $0.72$ & $3.71$ & $0.65$ & $0.59$ & $4.07$ & $0.63$ & $0.58$\\
     \hline
     
    $10$ & VLN$\circlearrowright$BERT~\cite{hong2021vln} & $2.90$ & $0.72$ & $0.68$ & $3.93$ & $0.63$ & $0.57$ & $4.09$ & $0.63$ & $0.57$\\
    $11$ &  VLN$\circlearrowright$BERT$^+$\textit{(ours)} & $2.72$ & $0.75$ & $0.70$ & $3.65$ & $0.65$ & $0.60$ & $4.09$ & $0.63$ & $0.57$\\
    $12$ &  VLN$\circlearrowright$BERT$^{++}$ \textit{(ours)} & $2.51$ & $0.77$ & $0.72$ & $3.40$ & $0.67$ & $0.61$ & $4.02$ & $0.63$ & $0.58$\\
    $13$ & VLN-Trans-R2R \textit{(ours)} & $\mathbf{2.40}$ &  $\mathbf{0.78}$ & $\mathbf{0.73}$ & $3.37$ & $0.67$ & $\mathbf{0.63}$ & $3.94$ & $0.65$ & $0.59$ \\
    $14$ & VLN-Trans-FG-R2R \textit{(ours)} & $2.45$ &  $0.77$ & $0.72$ & $\mathbf{3.34}$ & $\mathbf{0.69}$ & $\mathbf{0.63}$ & $3.94$ & $\mathbf{0.66}$ & $\mathbf{0.60}$ \\
    
    \hline
    \end{tabular}
    }
    \end{center}
    \vspace{-4mm}
    \caption{\small Experimental results on R2R Benchmarks in a single-run setting. The best results are in bold font. + means we add RXR~\cite{ku2020room} and Marky-mT5 dataset~\cite{wang2022less} as the extra data to pre-train the navigation agent. ++ means we further add SyFiS dataset to pre-train the navigation agent. ViT means Vision Transformer representations.}
    \label{R2R Experimental Result}
       \vspace{-2mm}
\end{table*}

\begin{table*}[!t]
\small
\renewcommand{\tabcolsep}{0.35em}
    \begin{center}
    \resizebox{1.0\textwidth}{!}{
    \begin{tabular}{ c c c c c c c c c c c c}
    \hline
    & & \multicolumn{5}{c}{\textbf{Val Seen}} & \multicolumn{5}{c}{\textbf{Val Unseen}} \\
    \hline
    & Method & \textbf{NE}$\uparrow$ & \textbf{SR}$\uparrow$ &  \textbf{SPL}$\uparrow$ & \textbf{CLS}$\uparrow$ & \textbf{sDTW}$\uparrow$ & \textbf{NE}$\downarrow$ & \textbf{SR}$\uparrow$ & \textbf{SPL}$\uparrow$ & \textbf{CLS}$\uparrow$ & \textbf{sDTW}$\uparrow$  \\
    \hline
    $1$ & OAAM~\cite{qi2020object} & - & $0.56$ & $0.49$ & $0.54$ & - & $0.32$  & $0.29$ & $0.18$ & $0.34$  & $0.11$\\
    $2$ & RelGraph~\cite{hong2020language}  & $5.14$ & $0.55$ & $0.50$ & $0.51$ &  $0.35$ & $7.55$ & $0.35$ & $0.25$ & $0.37$  & $0.18$\\
    $3$ & NvEM~\cite{an2021neighbor}  & $5.38$ & $0.54$ & $0.47$ & $0.51$ &  $0.35$ & $6.80$ & $0.38$ & $0.28$ & $0.41$ & $0.20$\\
    \hline
    $4$ & VLN$\circlearrowright$BERT*~\cite{hong2021vln}  & $4.82$ & $0.56$ & $0.46$ & $0.56$ & $0.38$ & $6.48$ & $0.43$ & $0.32$ & $0.42$ & $0.21$\\
    $5$ & CITL~\cite{liang2022contrastive} & $\mathbf{3.48}$ & $0.67$ & $0.57$ & $0.56$ & $\mathbf{0.43}$ & $6.42$ & $0.44$ & $0.35$ & $0.39$ & $0.23$ \\
    $6$ & LOViS~\cite{zhang2022lovis} & $4.16$ & $\mathbf{0.67}$ & $0.58$  & $\mathbf{0.58}$ & $\mathbf{0.43}$ & $6.07$ & $0.45$ & $0.35$ & $\mathbf{0.45}$ & $0.23$\\
    \hline
    $7$ & VLN-Trans & $3.79$ & $\mathbf{0.67}$  & $\mathbf{0.59}$ &  $0.57$ & $\mathbf{0.43}$  &  $\mathbf{5.87}$ & $\mathbf{0.46}$ & $\mathbf{0.36}$ & $\mathbf{0.45}$ & $\mathbf{0.25}$\\
    \hline
    \end{tabular}
    }
    \end{center}
    \vspace{-3mm}
    \caption{\small Experimental results on R4R dataset in a single-run setting. * denotes our reproduced R4R results.\vspace{-3mm}}
    \label{Experiment Results for R4R}
\end{table*}
\begin{table}[!t]
\small
\renewcommand{\tabcolsep}{0.35em}
    \begin{center}
    \resizebox{0.48\textwidth}{!}{
    \begin{tabular}{ c c c c c}
    \hline
    & \multicolumn{2}{c}{\textbf{Val Seen}} & \multicolumn{2}{c}{\textbf{Val Unseen}} \\
    \hline
    Method  & \textbf{SR}$\uparrow$ & \textbf{SPL}$\uparrow$ & \textbf{SR}$\uparrow$ & \textbf{SPL}$\uparrow$ \\
    \hline
    EnvDrop~\cite{tan2019learning} & $0.43$ & $0.38$ & $0.34$ & $0.28$ \\
    VLN$\circlearrowright$BERT~\cite{hong2020language} & $0.50$ & $0.46$ & $0.42$ & $0.37$ \\
    HAMT~\cite{chen2021history} & $0.53$ & $0.50$ & $0.45$ & $0.41$ \\
    VLN-Trans & $\mathbf{0.58}$ & $\mathbf{0.53}$ & $\mathbf{0.50}$ & $\mathbf{0.45}$ \\
    \hline
    \end{tabular}
    }
    \end{center}
    
    \vspace{-3mm}
    \caption{\small Experimental results on the R2R-Last dataset. \vspace{-4mm}}
    \label{Experiment Results for R2R-Last}
\end{table}

\subsection{Dataset and Evaluation Metrics}
\noindent\textbf{Dataset} We evaluate our approach on three datasets: \textbf{R2R}~\cite{anderson2018vision}, \textbf{R4R}~\cite{jain2019stay}, and \textbf{R2R-Last}~\cite{chen2021history}. R2R includes $21,567$ instructions and $7,198$ paths. The entire dataset is partitioned into training, seen validation and unseen validation, and unseen test sets. R4R extends R2R with longer instructions by concatenating two adjacent tail-to-head  trajectories in R2R. R2R-Last uses the last sentence in the original R2R to describe the final destination instead of step-by-step instructions. 

\noindent\textbf{Evaluation Metrics} 
Three metrics are used for navigation~\cite{anderson2018vision}:(1) Navigation Error (NE): the mean of the shortest path distance between the agent's final position and the goal destination. (2) Success Rate (SR): the percentage 
of the predicted final position being within $3$ meters from the goal destination. (3) Success rate weighted Path Length (SPL) that normalizes the success rate with trajectory length.
The R4R dataset uses two more metrics
to measure the fidelity between the predicted and the ground-truth path:
(4) Coverage Weighted by Length Score~(CLS)~\cite{jain2019stay}. (5) Normalized Dynamic Time Warping weighted by Success Rate~(sDTW)~\cite{ilharco2019general}.
We provide a more detailed description of the dataset and metrics in the Appendix~\ref{appendix: evlautation}.

\subsection{Implementation Details}
We use ResNet-152~\cite{he2016deep} pre-trained on Places365~\cite{zhou2017places} as the visual feature and the pre-trained BERT~\cite{vaswani2017attention} representation as the initialized text feature. 
We first pre-train the translator and navigation agent offline. Then we include the translator in the navigation agent to train together.
To pre-train the translator, we use one NVIDIA RTX GPU. 
The batch size and learning rate are $16$ and $1e-5$, respectively. Both $\alpha_1$ and $\alpha_2$ in Eq.~\ref{Eq_pretrain} are $1$. 
To pre-train the navigation agent, we follow the methods in  \newcite{zhang2022lovis} and use extra pre-training datasets to improve the baseline. We use $4$ GeForce RTX $2080$ GPUs, and the batch size on each GPU is $28$.The learning rate is $5e-5$.

We further train the navigation agent with a translator for $300$K iterations using an NVIDIA RTX GPU. The batch size is $16$, and the learning rate is $1e-5$. The optimizer is AdamW~\cite{loshchilov2017decoupled}. We can get the best results when we set  $\lambda$ as $0.2$ in Eq.~\ref{equation:nav} , and $\beta_1$, $\beta_2$, and $\beta_3$ as $1$, $1$ and $0.1$ in Eq.~\ref{Eq_finetune}, respectively. Please check our code~\footnote{\url{https://github.com/HLR/VLN-trans}} for the implementation.

\subsection{Experimental Results}
Table~\ref{R2R Experimental Result} shows the model performance on the R2R benchmarks. Row \#4 to row \#9 are Transformer-based navigation agents with pre-trained cross-modality representations, and such representations greatly improve performance of LSTM-based VLN models
(row \#1 to row \#3). 
It is impressive that our VLN-Trans model's performance (row \#13 and row \#14) on both validation seen and unseen performs $2\%$-$3\%$ better than HAMT~\cite{chen2021history} when it even uses more advanced ViT~\cite{dosovitskiy2020image} visual representations compared with ResNet.
Our performance on both SR and SPL are still $3\%$-$4\%$ better than the VLN agent using contrastive learning: CITL~\cite{liang2022contrastive} (row \#7) and ADAPT~\cite{lin2022adapt} (row \#8). LOViS~\cite{zhang2022lovis} (row \#9) is another very recent SOTA improving the pre-training representations of the navigation agent, but we can significantly surpass their performance.
Lastly, compared to the baseline (row \#10), we first significantly improve the performance (row \#11) by using extra augmented data, Room-across-Room dataset (RXR)~\cite{ku2020room} and the Marky-mT5~\cite{wang2022less}, in the pre-training of navigation agent.
The performance continues to improve when we further include the SyFiS dataset in the pre-training, as shown in row \#12, proving the effectiveness of our synthetic data.
Row \#13 and row \#14 are the experimental results after incorporating our pre-trained translator into the navigation model.
First, for a fair comparison with other models, we follow the baseline~\cite{hong2021vln} to train the navigation agent using the R2R~\cite{anderson2018vision} dataset and the augmented data from PREVALENT~\cite{hao2020towards}.
Since those datasets only contain the pairs of full instructions and the trajectories without intermediate alignments between sub-instructions and the corresponding viewpoints, we do not optimize the translator ($\beta2=0$, $\beta3=0$ in Eq.\ref{Eq_finetune}) during training the navigation agent, which is denoted as VLN-Trans-R2R. As shown in row \#13, our translator helps the navigation agent obtain the best results on the seen environment and improves SPL by $2\%$ on the unseen validation environment, proving that the generated sub-instruction representation enhances the model's generalizability.
However, FG-R2R~\cite{hong2020sub} provides human-annotated alignments between sub-instructions and viewpoints for the R2R dataset, and our SyFiS dataset also provides synthetic sub-instructions for each viewpoint. 
Then we conduct another experiment using FG-R2R and SyFiS datasets to train the navigation agent. Simultaneously, we optimize the translator using the alignment information with our designed SIG and SS losses during the navigation process.
As shown in row \#13, we further improve the SR and SPL on the unseen validation environment. This result indicates our designed losses can better utilize the alignment information.


Table~\ref{Experiment Results for R4R} shows results on the R4R benchmark. 
Row \#1 to Row \#3 are the LSTM-based navigation agent. Row \#4 reports our re-implemented results of VLN$\circlearrowright$BERT, and both 
CITL and LOViS are the SOTA models.
Our method (row \#7) improves the performance on almost all evaluation metrics, especially in the unseen environment. The high sDTW means that our method helps navigation agents reach the destination with a higher successful rate and better follow the instruction.

Table~\ref{Experiment Results for R2R-Last} shows the performance on the R2R-Last benchmark.
When only the last sub-sentence is available, 
our translator can generate a sub-instruction representation that assists the agent in approaching the destination. As shown in Table~\ref{Experiment Results for R2R-Last}, we improve the SOTA (Row \#3) by almost $5\%$ on the SR in the unseen validation dataset. We obtain the best results on R2R-Last without the Sub-instruction Split task. More details are in the ablation study (see Sec.~\ref{ablation study}). 
\begin{table}[!t]
\renewcommand{\tabcolsep}{0.45em}
    \begin{center}
    \resizebox{0.49\textwidth}{!}{
    \begin{tabular}{c|c|ccc|cc|ccc}
    \hline
     \multirow{2}{*}{\textbf{Dataset}}&\multirow{2}{*}{\textbf{Method}}& \multicolumn{3}{c}{\textbf{Tasks}} & \multicolumn{2}{|c}{\textbf{Val Seen}} & \multicolumn{2}{|c}{\textbf{Val Unseen}} \\
     && \textbf{SIG} & \textbf{DSL} & \textbf{SS} & \textbf{SR}$\uparrow$ & \textbf{SPL}$\uparrow$ & \textbf{SR}$\uparrow$ & \textbf{SPL}$\uparrow$ \\
    \hline
    \multirow{4}{*}{R2R}& Baseline & & & & $0.767$ & $0.722$ & $0.672$ & $0.611$ \\
    & $1$ & \ding{52} & & &  $0.764$ & $0.721$ & $0.673$ & $0.623$ \\
    & $2$ &  \ding{52} & \ding{52} & & $\mathbf{0.780}$ & $\mathbf{0.728}$ & $0.674$ & $0.627$ \\
    & $3$ & \ding{52}& \ding{52} & \ding{52}& $0.772$ & $0.720$ & $\mathbf{0.690}$ & $\mathbf{0.633}$ \\
    \hline
    \multirow{4}{*}{R2R-Last}& Baseline & & & & $0.552$ & $0.501$ & $0.473$ & $0.422$ \\
    & $1$ & \ding{52} & & &  $0.573$ & $0.521$ & $0.494$ & $0.434$ \\
    & $2$ &  \ding{52} & \ding{52} & & $\mathbf{0.582}$ & $\mathbf{0.534}$ & $\mathbf{0.503}$ & $\mathbf{0.453}$ \\
    & $3$ & \ding{52} & \ding{52} & \ding{52} & $0.571$ & $0.511$ & $0.484$ & $0.433$\\
    \hline
    \end{tabular}
    }
    \end{center}
    \caption{\small Ablation study, where Baseline is VLN$\circlearrowright$BERT$^{++}$. }
    \label{Ablation study on R2R}
    \vspace{-5.5mm}
\end{table}
\subsection{Ablation Study}
\label{ablation study}
In Table~\ref{Ablation study on R2R},
we show the performance after ablating different tasks in the baseline model on the R2R and R2R-Last datasets. We compared with VLN$\circlearrowright$BERT$^{++}$, which is our improved baseline after adding extra pre-training data to the navigation agent. 
First, we pre-train our translator with SIG and DSL tasks and incorporate the translator into the navigation agent without further training.
For both the R2R dataset and R2R-Last, SIG and DSL pre-training tasks can incrementally improve the unseen performance (as shown in method 1 and method 2 for R2R and R2R-Last).
Then we evaluate the effectiveness of the SS task when we use it to train the translator together with the navigation agent. For the R2R dataset, the model obtains the best result on the unseen environment after using the SS task. However, the SS task causes the performance drop for the R2R-Last dataset.
This is because the R2R-Last dataset merely has the last single sub-instruction in each example and there is no other sub-instructions our model can identify and learn from.

\subsection{Qualitative Study}
\noindent\textbf{Statistic of the SyFiS dataset} 
We construct SyFiS dataset using $1,076,818$ trajectories, where  $7198$ trajectories are from the R2R dataset, and $1,069,620$ trajectories are from the augmented data~\cite{hao2020towards}. Then we pair those trajectories with our synthetic instructions to construct the SyFiS dataset based on our pre-defined motion verb vocabulary and CLIP-generated landmarks (in Sec\ref{Synthetic Data}). When we pre-train the translator, we use the sub-instruction of each viewpoint in a trajectory. There are usually $5$ to $7$ viewpoints in a trajectory; each viewpoint is with one positive sub-instruction and three negative sub-instructions. 

\noindent\textbf{Quality of SyFiS dataset.} 
\label{qualitative study}
We randomly select $50$ instructions from the SyFiS dataset and manually check if humans can easily follow those instructions. 
As a result, we achieve $58\%$ 
success rate. It is reported~\cite{wang2022less} that success rate of the generated instruction are $38\%$ and $48\%$ in Speaker-Follower~\cite{fried2018speaker} and Env-dropout~\cite{tan2019learning}, respectively. The $10\%$ higher success rate of our instructions indicates we have synthesized a better quality dataset for pre-training and fine-tuning. 


\noindent\textbf{Translator Analysis} 
Our translator can relate the mentioned landmarks in the instruction to 
the visible and distinctive landmarks in the visual environment. 
In Fig.~\ref{qualitative study}~(a), 
``tables'' and ``chairs'' are not visible in three candidate viewpoints (v$1$-v$3$). However, our navigation agent can correctly recognize the target viewpoint using the implicit instruction representations generated by the translator.
We assume the most recognizable and distinctive landmark, that is, the "patio" here in the viewpoint v$3$ has a higher chance to be connected to a ``table'' and a ``chair'' based on our pre-training, compared to the landmarks in the other viewpoints.
In Fig.~\ref{qualitative study} (b), both candidate viewpoints v$2$ and v$3$ contain kitchen (green bounding boxes); hence it is hard to distinguish the target between them. However, for the translator, the most distinctive landmark in v$3$ is the ``cupboard'' which is 
more likely to be related to the ``kitchen''.
Fig.~\ref{qualitative study}(c) shows a failure case, in which the most distinctive landmark in candidate viewpoint v$1$ is ``oven''. It is more likely for the  translator relates ``oven'' to the ``kitchen'' compared to ``countertop'', and the agent selects the wrong viewpoints.
In fact, we observe that the R2R validation unseen dataset has around $300$ instructions containing ``kitchen''. For corresponding viewpoints paired with such instructions, our SyFiS dataset generates $23$ and $5$ sub-instructions containing  ``oven'' and ``countertop'', respectively, indicating the trained translator more likely relates ``oven'' to ``kitchen''. More examples are shown in Appendix.~\ref{Qualitative Examples for Translator Analysis}.
\begin{figure}
    \centering
     \begin{overpic}[width=1.\linewidth]{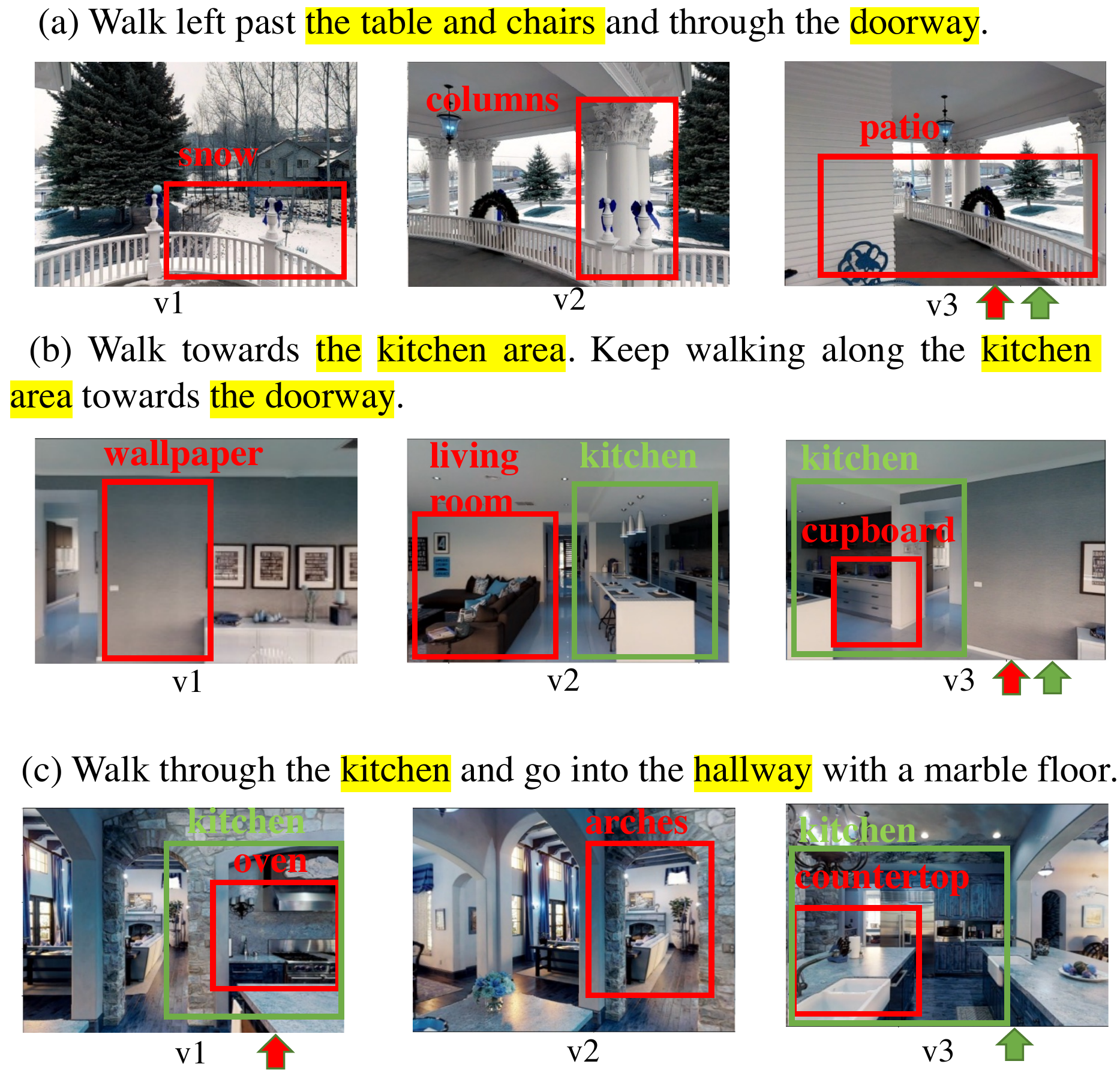}
    \end{overpic}
    \caption{Qualitative examples to show how the translator helps the navigation agent.
    The red boxes and green boxes show the distinctive and the nondistinctive landmarks; 
    the green arrow and red arrow show the target and the predicted viewpoints.}
    \label{qualitative study}
    \vspace{-5mm}
\end{figure}


\vspace{-1.5mm}
\section{Conclusion}
In the VLN task, instructions given to the agent often include landmarks that are not recognizable to the agent or are not distinctive enough to specify the target.
Our novel idea to solve these issues is to include a translator module in the navigation agent that converts the given instruction representations into effective sub-instruction representations at each navigation step. 
To train the translator, we construct a synthetic dataset and design pre-training tasks to encourage the translator to generate the sub-instruction with the most recognizable and distinctive landmarks. 
Our method achieves the SOTA results on multiple navigation datasets. We also provide a comprehensive analysis to show the effectiveness of our method. It is worth noting that while we focus on R2R, the novel components of our technique for generating synthetic data and pre-training the translator are easily applicable to other simulation environments.

\section{Limitations}
We mainly summarize three limitations of our work. First,  the translator only generates a representation, not an actual instruction, making the model less interpretable. Second, we do not include more advanced vision representations such as ViT and CLIP to train the navigation agent.
Although only using ResNet, we already surpass prior methods using those visual representations (e.g., HAMT~\cite{chen2021history}), it would be interesting to experiment with those different visual representations. Third, this navigation agent is trained in a simulated environment, and a more realistic setting will be more challenging.



\section{Acknowledgement}
This project is supported by National Science Foundation (NSF) CAREER award 2028626 and partially supported by the Office of Naval Research
(ONR) grant N00014-20-1-2005. Any opinions,
findings, and conclusions or recommendations expressed in this material are those of the authors and
do not necessarily reflect the views of the National
Science Foundation nor the Office of Naval Research. We thank all reviewers for their thoughtful
comments and suggestions.

\bibliography{custom}
\bibliographystyle{acl_natbib}

\clearpage
\appendix
\clearpage
\section{Appendix}
\subsection{Motion Indicator Dictionary}
\label{motion indicator dictionary}
\begin{figure}
    \centering
    \includegraphics[width=1.\linewidth]{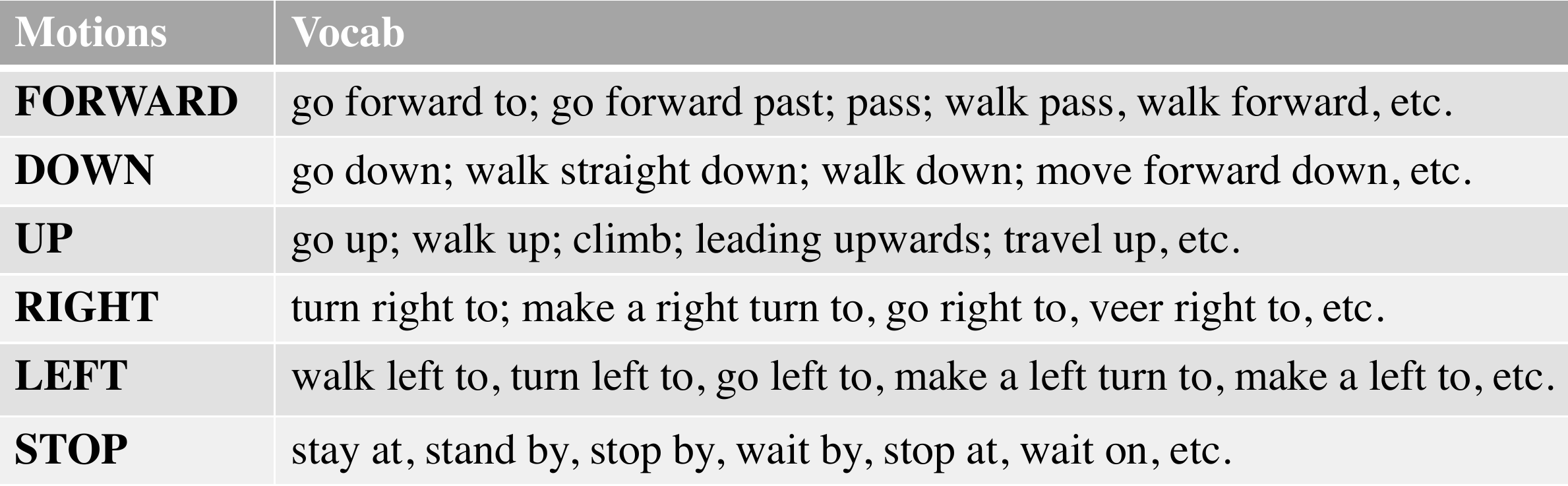}
    \caption{Motion Indicator Vocabulary}
    \label{fig:Vocabulary of the Motions}
    \vspace{-5mm}
\end{figure}
We extract the motion verb phrases in the R2R training instructions to build a motion indicator dictionary, as shown in Fig.~\ref{fig:Vocabulary of the Motions}. We first use spaCy~\footnote{\url{https://spacy.io/}} to extract motion verbs based on pos-tagging information , and then manually collect the prepositions after the motion verbs, such as ``stop at'', '' stop by'' and '' stop behind of''.
In summary, there are $131$ verb phrases for the action of ``FORWARD'', $11$ verb phrases for the action of ``DOWN'', $11$ verb phrases for the action of``UP'', $28$ verb phrases for the action of``LEFT'', $23$ for the action of ''RIGHT'', and $26$ for the action of ''STOP''.
\subsection{Comparison among different datasets}
\begin{figure}
    \centering
    \includegraphics[width=1.\linewidth]{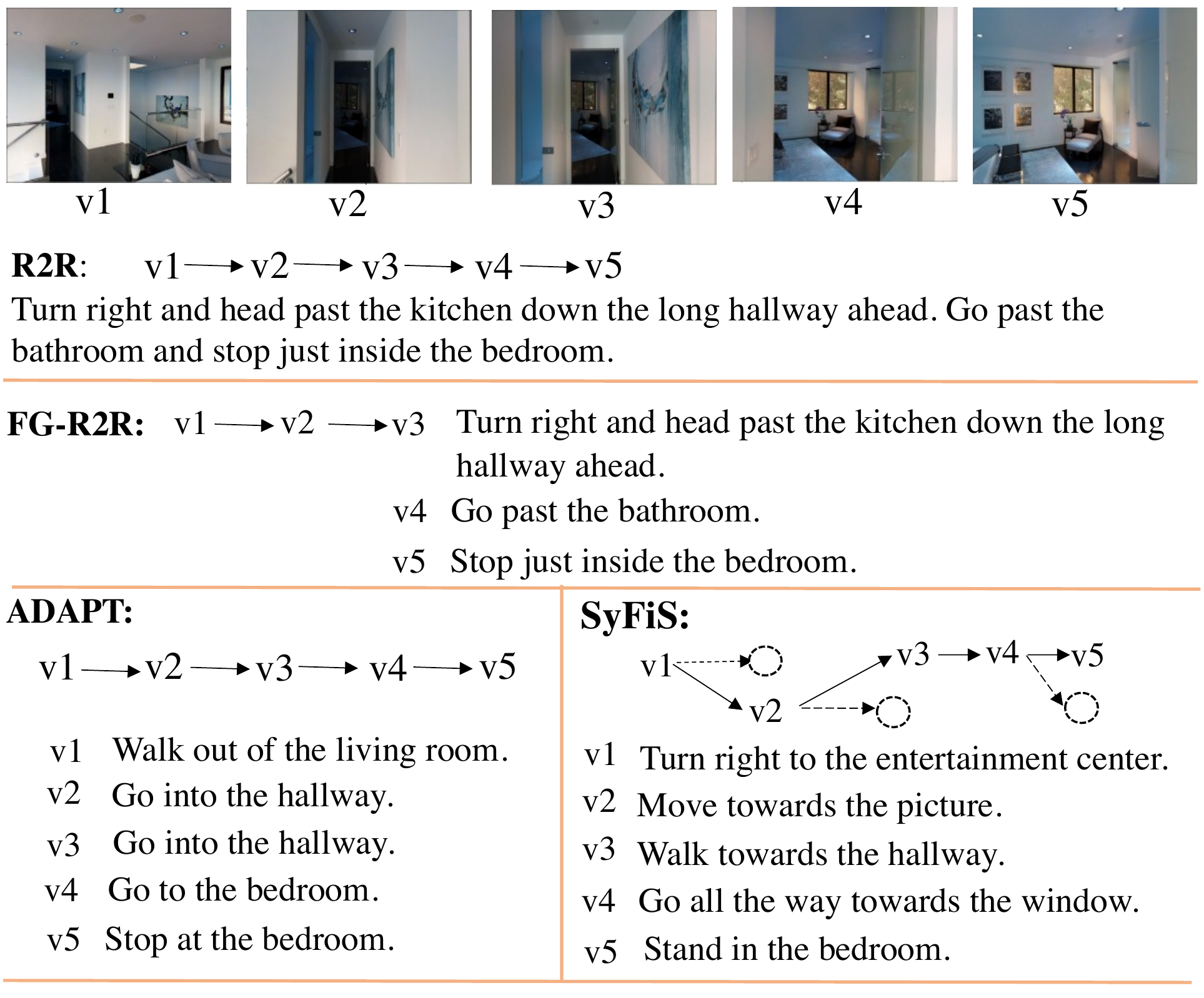}
    \caption{Comparisons among Different Datasets.}
    \label{fig:dataset comparision}
    \vspace{-5mm}
\end{figure}

One of the contributions of our method is the proposed SyFiS dataset, which forms sub-instruction for each viewpoint considering recognizable and distinguishable landmarks. In this section, we compare different datasets to show the main improvements of the SyFiS compared to other datasets. As shown in Fig.~\ref{fig:dataset comparision}, in the R2R dataset~\cite{anderson2018vision}, instructions describe the entire trajectory, which is challenging for the navigation agent to follow in every single step. Based on it, FG-R2R~\cite{hong2020sub} provides a manual annotation to align the sub-instruction to the corresponding viewpoints. Although providing fine-grained annotation, the sub-instructions in FG-R2R are still not step-by-step.
ADAPT~\cite{lin2022adapt} generates the sub-instruction for every single viewpoint. 
However, they only consider the viewpoints in trajectory and select the most obvious landmarks for each target viewpoint. 
Those selected landmarks are quite general, and hard to distinguish the target viewpoint from other candidate viewpoints, such as the ``living room'', ``hallway'' and ``bedroom''.
Nevertheless, both FG-R2R and ADAPT still suffer from the issue of nondistinctive landmarks, such as the ``living room'', ``hallway'' and ``bedroom'', which hurts the navigation performance, as stated previously. 
We construct a dataset with the most recognizable and distinguishable landmark, which is obtained by comparing the target viewpoint with other candidate viewpoints at each navigation step. Based on our experimental results, our generated sub-instruction dataset can largely help the navigation performance. 


\begin{figure}[htp]
\subfloat[\scriptsize{The landmark of the ``fireplace'' in the given instruction can not be observed in all three candidate viewpoints. The target viewpoint v3 contains a distinctive landmark ``living room'', and the translator can relate it to ``fireplace'', which helps the agent select the correct target viewpoint.}]{%
  \includegraphics[width=1.0\linewidth]{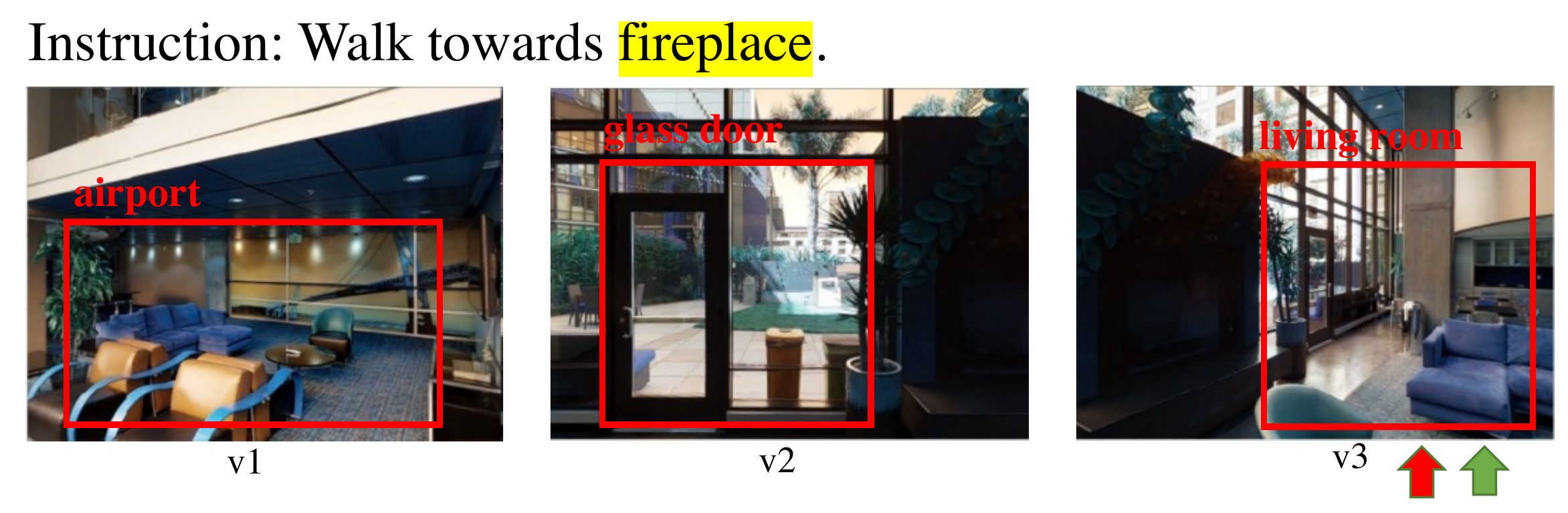}
}

\subfloat[\scriptsize{The target viewpoint v1 contains landmarks of ``apartment'', ``cupboard'' and ``skylight''. Among them, ``apartment'' and ``cupboard'' are nondistinctive because they appear in both target and other candidate viewpoints. Our agent can select the correct targe viewpoint because our translator can relate ``skylight`` to ``balcony``. According to our observations, in the R2R training data, among the trajectories paired with the instructions containing ``balcony'', our SyFiS dataset generates 14\% sub-instructions containing "skylight" for the corresponding viewpoints.}]{%
  \includegraphics[width=1.0\linewidth]{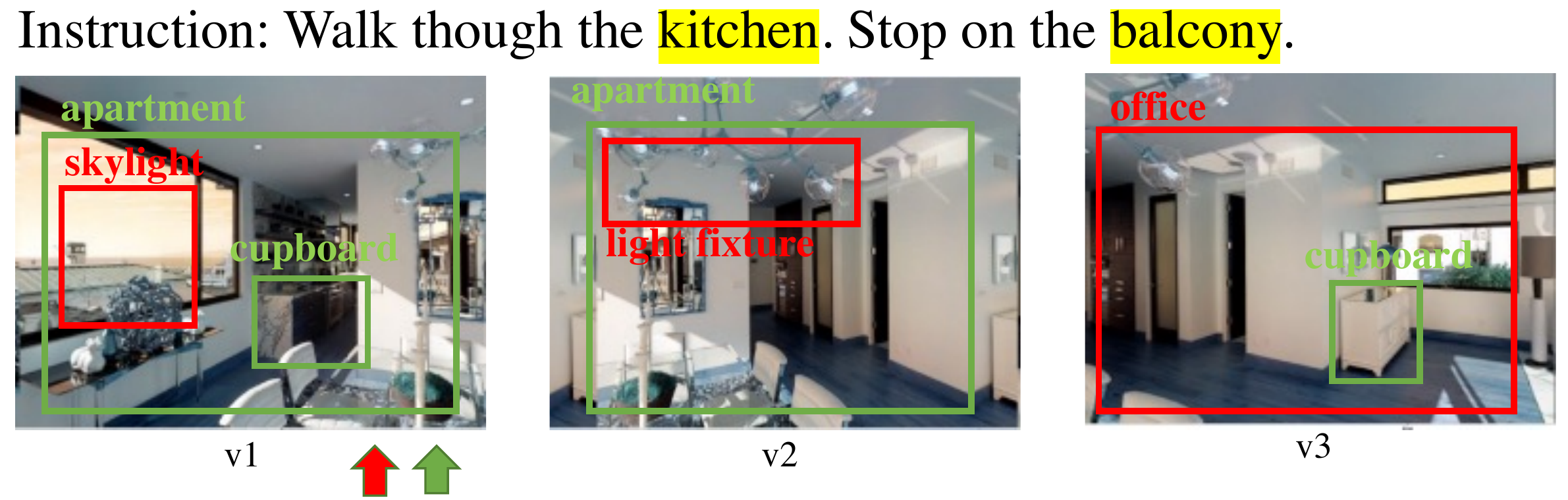}
}

\subfloat[\scriptsize{The landmarks of the ``bathroom'' in the instruction can not be observed in all viewpoints. The target viewpoint v1 contains landmarks of ``hallway'' and ``vanity'', where ``hallway'' is nondistinctive since it also can be observed in other candidate viewpoints. Our translator relates ``vanity'' to ``bathroom'', which helps the agent select the correct viewpoint.}]{
\includegraphics[width=1.0\linewidth]{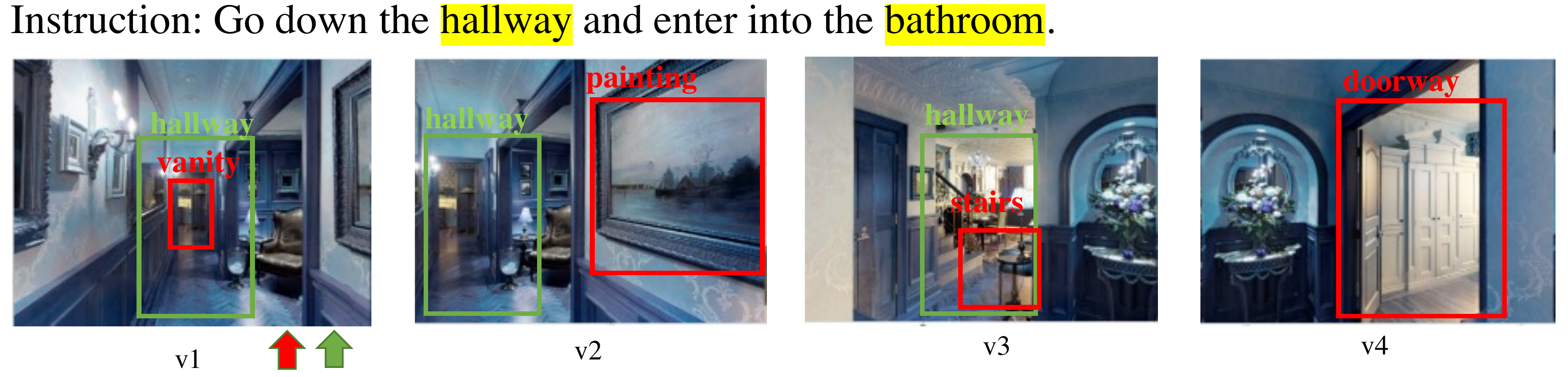}
}

\subfloat[\scriptsize{In this case, all candidate viewpoints include ``hall''. According to our observation, in the R2R training data, among the trajectories paired with the instructions containing ``hall'', our SyFiS dataset generates 3\% sub-instructions containing "pillar'', 1\% containing ``court'', and  almost 0\% containing ``church''.In such a case, our translator has a higher chance of relating ``hall'' to ``pillar'' and selecting the wrong viewpoint.}]{%
  \includegraphics[width=1.0\linewidth]{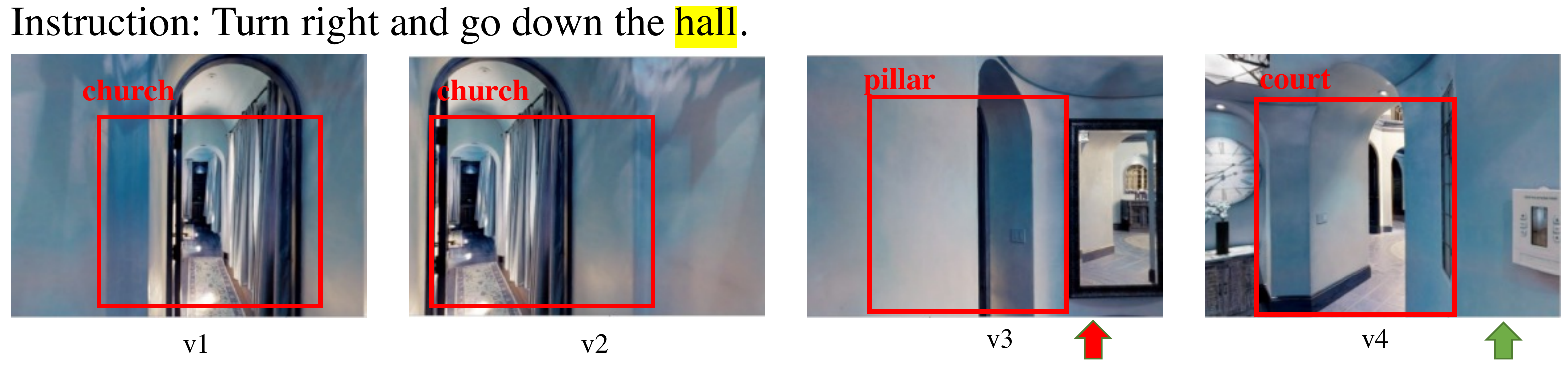}
}

\subfloat[\scriptsize{Both viewpoints contain ``kitchen'' and ``hall'', but our translator highly relates ``kitchen'' to ''cabinets'' compared to ``oven''. In the R2R training data, among the trajectories paired with the instructions containing ``kitchen'', our SyFiS dataset generates 9\% sub-instructions containing "cabinet'' while 6\% containing ``oven''. In such a case, our translator is more likely to relate ``kitchen'' to ``cabinets'' and select the wrong viewpoint.}]{%
  \includegraphics[width=0.8\linewidth]{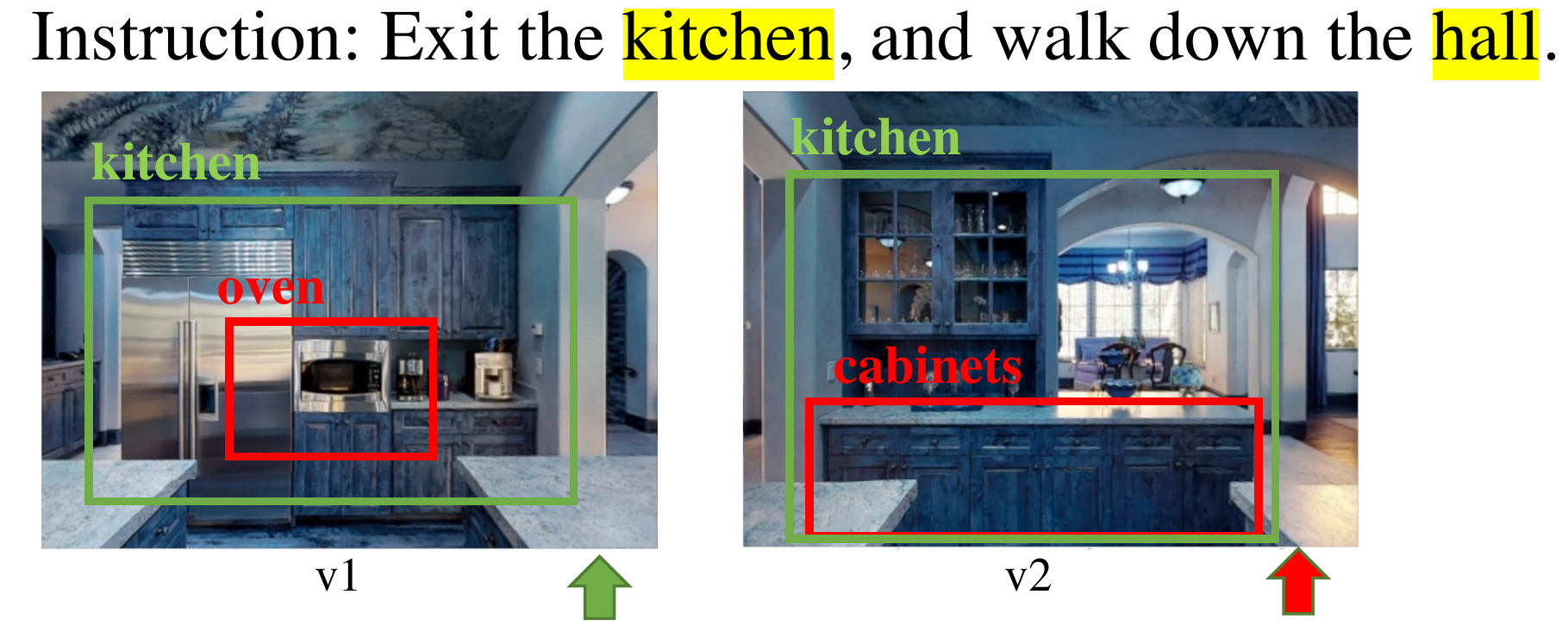}
}

\caption{Qualitative Examples. (a)(b)(c) are correct examples, and (d)(e) are wrong examples. The red boxes and green boxes show the distinctive and nondistinctive landmarks based on the target viewpoint; The green arrow and red arrow show the target and the predicted viewpoint from model.}
\label{qualitative example2}
\end{figure}

\subsection{Evaluation Datasets and Metrics} 
\label{appendix: evlautation}
Our method is evaluated on R2R~\cite{anderson2018vision}, R4R~\cite{jain2019stay}, and R2R-Last~\cite{chen2021history}. All these three dataset are built upon the Matterport3D~\cite{chang2017matterport3d} indoor scene dataset.\\
\textbf{R2R} provides long instructions paired with the corresponding trajectory. The dataset contains $61$ houses from training, $56$ houses for validation in seen environment, $11$ and $18$ houses for unseen environment validation and test, respectively. The seen set shares the same visual environment with training dataset, while unseen sets contain different environments. \\
\textbf{R4R} extends R2R by concatenating two trajectories and their corresponding instructions. In R4R, trajectories are less biased compared to R2R, because they are not necessarily the shortest path from the source viewpoint to the target viewpoint. \\
\textbf{R2R-Last} proposes a VLN setup that is similar to that of REVERIE~\cite{qi2020reverie}, which only claims the destination position.
More formally, R2R-Last only leverages the the last sentence in the original R2R instructions to describe the final destination.\\

\textbf{Evaluation Metrics}
VLN task mainly evaluates navigation agent's generalizability in unseen environment using validation unseen and test unseen datasets. Success Rate (SR) and Success Rate weighted Path length (SPL) are two main metrics for all three datasets, where a predicted path is \textit{success} if the agent stop within $3$ meters of the destination. The metrics of SR and SPL can evaluate the accuracy and efficiency of navigation.

\subsection{Qualitative Examples for Translator Analysis} 
\label{Qualitative Examples for Translator Analysis}
We provide more qualitative examples in Fig.~\ref{qualitative example2} to show our translator can relate the mentioned landmarks in the instruction to the recognizable and distinctive landmarks in the visual environment. 
Fig.~\ref{qualitative example2}(a)(b)(c) shows successful cases in that our translator helps the navigation agent make correct decisions.
However, there are chances our translator relates to wrong landmarks in the visual environment because of biased data. This may lead to the wrong decisions of the navigation agent, and we provide failure cases in Fig.~\ref{qualitative example2}(e)(f).


\end{document}